# ACTIVE SPLINE MODEL: A SHAPE BASED MODEL—INTERACTIVE SEGMENTATION

Jen Hong Tan[#], and U. Rajendra Acharya

*Abstract* - **Rarely in literature a method of segmentation cares for the edit after the algorithm delivers. They provide no solution when segmentation goes wrong. We propose to formulate point distribution model in terms of centripetal-parameterized Catmull-Rom spline. Such fusion brings interactivity to model-based segmentation, so that edit is better handled. When the delivered segment is unsatisfactory, user simply shifts points to vary the curve. We ran the method on three disparate imaging modalities and achieved an average overlap of 0.879 for automated lung segmentation on chest radiographs. The edit afterward improved the average overlap to 0.945, with a minimum of 0.925. The source code and the demo video are available at** http://wp.me/p3vCKy-2S

*Keywords* - Active spline models, model-based segmentation, point distribution model, Catmull-Rom spline

## 1. Introduction

In reality feasible segmentation demands more than accuracy [1,2]. Methods such as level set ([3,4] ), Markov random field ([5,6]) and other active strategies [7] can produce satisfactory segmentation, but to edit the segments afterward is never a simple feat. And although we can always complement any sophisticated segmentation with corrective measures that manipulate segment at pixel level or—in the case of contour-based method—the numerous points encircling the region of interest, such a strategy was considered rather insensible [8].

Therefore a homogeneous solution to segmentation and segment edit is desirable. This study proposes to fuse point distribution model [9] and centripetal Catmull-Rom spline (CRS) [10] to deal with the above necessity.

Point distribution model was devised to provide model specificity [9]. The method uses a set of points to denote salient features of an object, captures their variations by principal component analysis and forms the model. When the model is put under an image force field, it can only deform in ways that are characteristic of the objects it is built to form.

---

[#]contact: tanjenhong@gmail.com; website: http://jenh.co

In point distribution model straight line is used to join two consecutive points, thus in application the model often takes considerable amount of points just to form curves. This inevitably introduces redundancy in points, as a result majority of the points denote not salient features, but part of some curves.

And the redundancy has at least two undesirable consequences. First, it renders active shape model or active appearance model [11] unsuitable for segment edit. A simple edit in the methods often involves a shift in considerable amount of points. Second, it makes the model training laborious. Users are forced to put up oodles of points just to smooth out the curves in every training sample, and landmark points lose their characteristics as true landmarks. These issues can be rectified if only when a few points are enough to delineate a curve.

Hence, we propose to fuse point distribution model and spline. However, neither cubic Bézier spline curve [12] nor uniform B-spline [13]—two of the most used splines—joins the control points that define the curve. And although by recursive filtering [14] uniform B-spline can join points, it sometimes adds in unwanted salient features on the curve. If the method is used, the shape of the object will be corrupted.

We find that CRS is the better and simpler solution. It interpolates and thus joins control points, and is fast to reproduce. Furthermore, by centripetal parameterization the spline guarantees no cusp or self-intersections within a curve segment [10]. The curve thus tightly goes after points of interest, and when edit is necessary, a shift of points is all it needs.

We name our method active spline model (ASPLM). Since, we use spline to eliminate point redundancy in point distribution model. Besides, although we use gradient vector flow field [15] as the default external force field for deformation, the model gets update in a manner dissimilar to the ways of active shape model and active contour model [16]. The term spline is used to illustrate these differences.

The paper is structured in the below fashion. Section 2 details the mathematics to construct a CRS with centripetal parameterization, to align shapes, to build point distribution model under spline, and the steps to proceed the segmentation. For section 3, we provide 2 examples of the use of our algorithm on eye related images, and another on posterior-anterior chest radiographs, which is an automated segmentation. We shall also illustrate how easy and precise the edit of segment can be after the segmentation. Finally, the pros and cons of the methods are discussed in section 4 and the paper concludes in section 5.

## 2. Method
### 2.1 Catmull-Rom spline with centripetal parameterization

Let $\mathbf{p} = [x \ y]^T$ denote a point. For a curve segment $\mathbf{Q}_i$ defined by control points $\mathbf{p}_{i-1}$, $\mathbf{p}_i$, $\mathbf{p}_{i+1}$, $\mathbf{p}_{i+2}$ and knot sequence $t_{i-1}, t_i, t_{i+1}, t_{i+2}$, we plot CRS [17] by

$$\mathbf{Q}_i = \frac{t_{i+1}-t}{t_{i+1}-t_i}\mathbf{L_{012}} + \frac{t-t_i}{t_{i+1}-t_i}\mathbf{L_{123}}$$

(1)

where

$$\mathbf{L_{012}} = \frac{t_{i+1}-t}{t_{i+1}-t_{i-1}}\mathbf{L_{01}} + \frac{t-t_{i-1}}{t_{i+1}-t_{i-1}}\mathbf{L_{12}}$$

$$\mathbf{L_{123}} = \frac{t_{i+2}-t}{t_{i+2}-t_i}\mathbf{L_{12}} + \frac{t-t_i}{t_{i+2}-t_i}\mathbf{L_{23}}$$

$$\mathbf{L_{01}} = \frac{t_i-t}{t_i-t_{i-1}}\mathbf{p}_{i-1} + \frac{t-t_{i-1}}{t_i-t_{i-1}}\mathbf{p}_i$$

$$\mathbf{L_{12}} = \frac{t_{i+1}-t}{t_{i+1}-t_i}\mathbf{p}_i + \frac{t-t_i}{t_{i+1}-t_i}\mathbf{p}_{i+1}$$

$$\mathbf{L_{23}} = \frac{t_{i+2}-t}{t_{i+2}-t_{i+1}}\mathbf{p}_{i+1} + \frac{t-t_{i+1}}{t_{i+2}-t_{i+1}}\mathbf{p}_{i+2}$$

and

(2)

$$t_{i+1} = |\mathbf{p}_{i+1} - \mathbf{p}_i|^\alpha + t_i$$

$\alpha$ ranges between 0 and 1, used for knot parameterization.

Suppose we want to draw a curve tightly going after $\mathbf{p}_1$, $\mathbf{p}_2$, $\mathbf{p}_3$, ... $\mathbf{p}_m$ by equations (1) and (2), however, only to realize the curve joins only from points $\mathbf{p}_2$ to $\mathbf{p}_{m-1}$. In fact, a curve segment $\mathbf{Q}_i$ defined by $\mathbf{p}_{i-1}$, $\mathbf{p}_i$, $\mathbf{p}_{i+1}$, $\mathbf{p}_{i+2}$ goes through only $\mathbf{p}_i$ and $\mathbf{p}_{i+1}$, not all the four. To rectify the shortcoming, we add in $\mathbf{p}_0$ and $\mathbf{p}_{m+1}$ before and after the point series:

(3)

$$\mathbf{p}_0 = \mathbf{p}_1 - \rho(\mathbf{p}_2 - \mathbf{p}_1)$$

(4)

$$\mathbf{p}_{m+1} = \mathbf{p}_m - \rho(\mathbf{p}_m - \mathbf{p}_{m-1})$$

where $\rho$ can be any value between 0 and 0.5, together with equations (1) and (2), let $i = 0,1,2,\dots,m+1$, $t_0 = 0$, the curve now goes through $\mathbf{p}_1$ to $\mathbf{p}_m$ (see Fig. 1).

On the other hand, when parameterization is not necessary, we can draw the spline simply by the original CRS [18]:

(5)

$$\mathbf{Q}_i = \frac{1}{2}\begin{bmatrix} t^3 \\ t^2 \\ t \\ 1 \end{bmatrix}^T \begin{bmatrix} -1 & 3 & -3 & 1 \\ 2 & -5 & 4 & -1 \\ -1 & 0 & 1 & 0 \\ 0 & 2 & 0 & 0 \end{bmatrix}\begin{bmatrix} \mathbf{p}_{i-3} \\ \mathbf{p}_{i-2} \\ \mathbf{p}_{i-1} \\ \mathbf{p}_i \end{bmatrix}$$

where $0 \le t \le 1$. The curve produced by equation (5) is equivalent to the one delineated by equations (1)—(4) with $\alpha$ equals to 0—a setup that generates a uniformly parameterized curve.

However, such configuration (or equation (5)) is never used in this work. As illustrated in Fig. 2 and 3, the cusps or self-intersections of the curve as a consequence of the parameterization is undesirable. Nor chordal parameterization—when $\alpha$ is set to 1—can fend off the woes. Only centripetal parameterization—when $\alpha$ equal to 0.5—produces spline of no cusp or self-intersection within a *curve segment*. This was mathematically proved by Yuksel *et al.* [10], and is thus consistently used throughout in this paper.

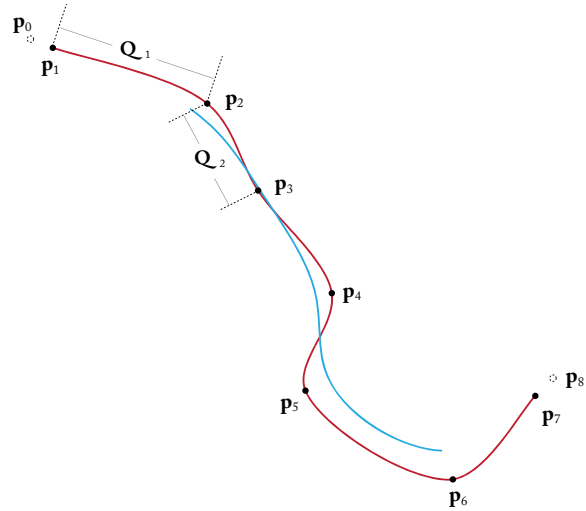

**Fig. 1.** Catmull-Rom spline (red) generated by equations (1)—(4) with centripetal parameterization, and also basis spline (blue), on the shape of *Valacirca*.

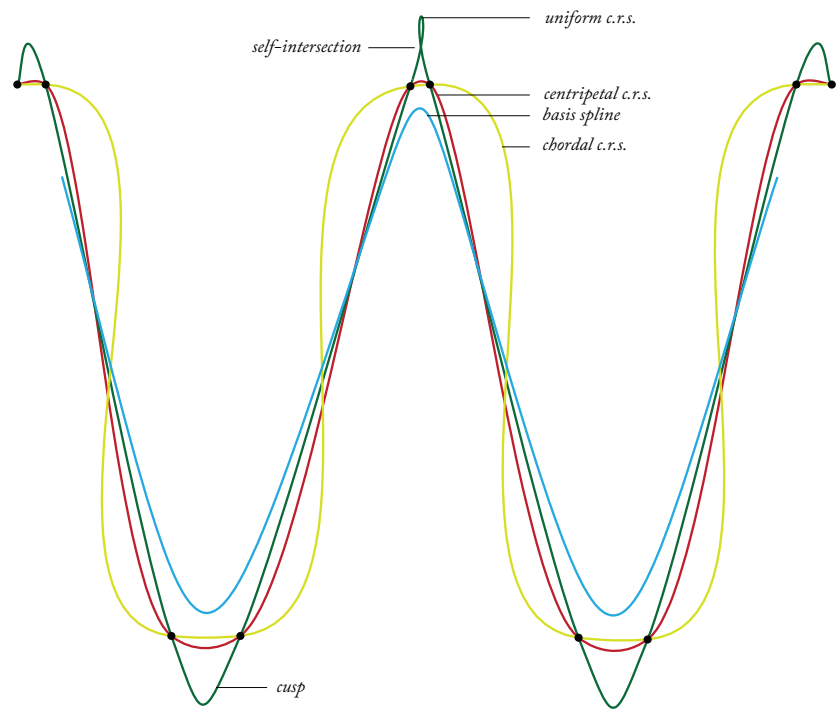

**Fig. 2.** Catmull-Rom splines by centripetal (red), uniform (green), chordal (yellow) parameterization, and basis spline (blue), on the shape of 'w'. 'c. r. s.' denotes Catmull-Rom spline.

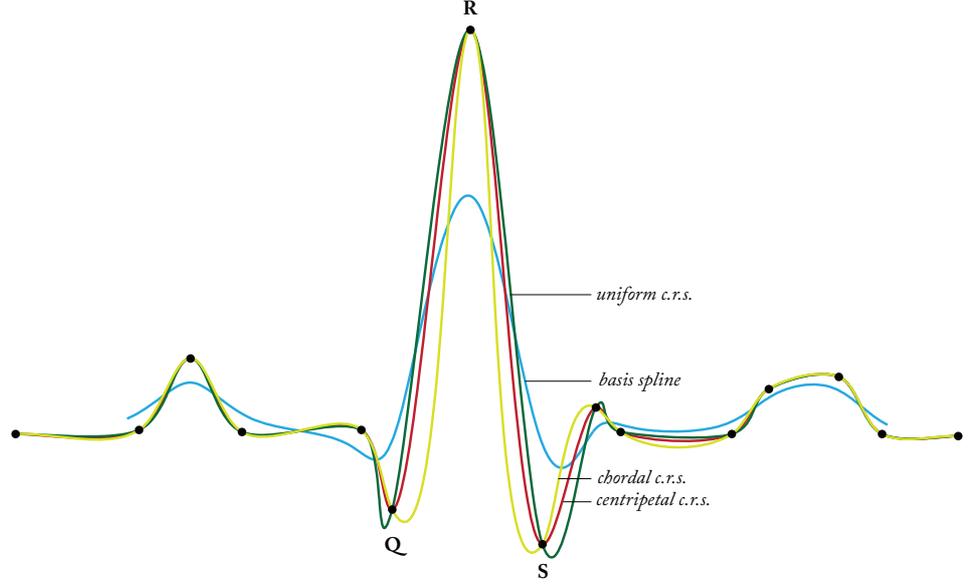

**Fig. 3.** Catmull-Rom splines by centripetal (red), uniform (green), chordal (yellow) parameterization, and basis spline (blue), on the shape of a typical electrocardiogram. At 'Q' and 'S', the peaks of the curves by uniform and chordal parameterization stray away of the intended locations. 'c. r. s.' denotes Catmull-Rom spline.

### 2.2 Pre-processes before formation of point distribution model

#### 2.2.1 Shape normalization

Let $\mathbf{C} = [x_1 \ y_1 \ x_2 \ y_2 \ ... \ x_i \ y_i \ ... \ x_m \ y_m]^T$ denote a shape of $m$ salient points where $\mathbf{p}_1 = [x_1 \ y_1]^T$, $\mathbf{p}_2 = [x_2 \ y_2]^T$, ... $\mathbf{p}_m = [x_m \ y_m]^T$, we normalize the shape by

$$\begin{cases} \hat{x}_i = \frac{x_i - \bar{x}}{\eta} \\ \hat{y}_i = \frac{y_i - \bar{y}}{\eta} \end{cases} \tag{6}$$

where

$$\eta = \max \left( \max_i \|x_i - \bar{x}\|, \max_i \|y_i - \bar{y}\| \right)$$

$\|\cdot\|$ refers to absolute value, and $\bar{x}$ and $\bar{y}$ denote the mean over $x_i$ and $y_i$ respectively for m salient points, so that each point $\hat{\mathbf{p}}_i = [\hat{x}_i \ \hat{y}_i]^T$ falls in the range of $0 \leq \hat{x}_i, \hat{y}_i \leq 1$.

#### 2.2.2 Affine transformation

For a shape $\mathbf{C}$ we can scale, rotate and translate the shape by application of the below transformation to each of the point in $\mathbf{C}$

$$\begin{cases} x' = \tau_x + x \cdot s \cos \theta - y \cdot s \sin \theta \\ y' = \tau_y + x \cdot s \sin \theta + y \cdot s \cos \theta \end{cases} \tag{7}$$

where $(x', y')$ is the point after the transformation, $\tau_x$, $\tau_y$ refer to the translation in $x$-axis and $y$-axis respectively; $s$ controls the scale and $\theta$ rotates the shape.

### 2.2.3 Shape alignment

Before the formation of point distribution model, it is necessary to align the shapes by (7). Given two shapes of the same ilk, $\mathbf{C}_{k-1}$ and $\mathbf{C}_k$, which have $m$ salient points, one need to determine the $\theta$, $s$, $\tau_x$, $\tau_y$ so that $\mathbf{C}_k$ is aligned to $\mathbf{C}_{k-1}$ when the transformation is applied (on $\mathbf{C}_k$)[9]:

$$
\begin{pmatrix}
\chi_k & -\gamma_k & W & 0 \\
\gamma_k & \chi_k & 0 & W \\
Z & 0 & \chi_k & \gamma_k \\
0 & Z & -\gamma_k & \chi_k
\end{pmatrix}
\begin{pmatrix}
a_x \\
a_y \\
\tau_x \\
\tau_y
\end{pmatrix}
=
\begin{pmatrix}
\chi_{k-1} \\
\gamma_{k-1} \\
C_1 \\
C_2
\end{pmatrix}
\tag{8}
$$

where

$$a_x = s\cos\theta \quad , \quad a_y = s\sin\theta$$

$$\chi_k = \sum_{i=1}^{m} w_i \, (x_i)_k \quad , \quad \gamma_k = \sum_{i=1}^{m} w_i \, (y_i)_k$$

$$Z = \sum_{i=1}^{m} w_i \left[ (x_i)_k^2 + (y_i)_k^2 \right] \quad , \quad W = \sum_{i=1}^{m} w_i$$

$$C_1 = \sum_{i=1}^{m} w_i \left[ (x_i)_{k-1} (x_i)_k + (y_i)_{k-1} (y_i)_k \right]$$

$$C_2 = \sum_{i=1}^{m} w_i \left[ (y_i)_{k-1} (x_i)_k - (x_i)_{k-1} (y_i)_k \right]$$

in which $[(x_i)_k \quad (y_i)_k]$ is the $i$th point in $\mathbf{C}_k$. Assume in total there are $r$ number of shapes in $\mathfrak{I}$ to be aligned, then $w_i$ in (8) is the weight for each pair of $(x_i)_k$ and $(y_i)_k$ in $\mathbf{C}_k$, imposed in the process of alignment to put emphasis on points that are consistent in their positions relative to the other peers across $S$. Let $(\delta_{ij})_k$ be the straight line distance between points $i$ and $j$ in $\mathbf{C}_k$, and $\sigma^2$ the variance of $(\delta_{ij})_k$ across the $\mathfrak{I}$, $w_i$ is deduced by

$$w_i = \frac{\omega_i}{\sum_{i=1}^{m} \omega_i} \tag{9}$$

$$\omega_i = \left( \sum_{j=1}^{m} \sigma_{\delta_{ij}}^2 \right)^{-1} \tag{10}$$

With these, for complete alignment of $r$ shapes in $\mathfrak{I}$, we do

---

Take any shape in $\mathfrak{I}$ as the align basis, and align every shape of $\mathfrak{I}$ with the basis.
**Repeat**
    Deduce the mean shape from all the aligned shapes.
    Align every shape in $\mathfrak{I}$ with the new mean shape.
**Until** the process converges.

---

### 2.2.4 Shape expansion

At times in the application of the method we may need to expand a shape by insertion of slave points after every master point $\mathbf{p}_i = [x_i \; y_i]$ (control point), so that the shape is well-behaved when put under image forces. These added points later can simply be discarded after segmentation with no adverse effect on delineation. Let $\mathbf{C} = [\mathbf{p}_1 \; \mathbf{p}_2 \; \dots \; \mathbf{p}_i \; \dots \; \mathbf{p}_m]^T$ denote the shape/spline before expansion, controlled by knot sequence $t_0, t_1, t_2, \dots t_i \dots t_{m+1}$ according to equations (1)—(4). Assume we want to insert $\epsilon$ number of slave points after every master point except the last, between $\mathbf{p}_i$ and $\mathbf{p}_{i+1}$, the slave points are generated by values of $t$ that equally sections the interval $t_{i+1} - t_i$.

Figure 4 illustrates 2 examples of expansion on shape 'S', $\mathbf{C} = [\mathbf{p}_1 \; \mathbf{p}_2 \; \dots \; \mathbf{p}_i \; \dots \; \mathbf{p}_7]^T$ with knot sequence $t_0, t_1, t_2, \dots t_i \dots t_8$. In Fig. 4(a), one slave point is inserted every after a master point (except the last), thus $\mathbf{s}^i$ is derived from $t$ set as $\frac{1}{2}(t_{i+1} - t_i)$ on points $\mathbf{p}_{i-1}, \mathbf{p}_i, \mathbf{p}_{i+1}, \mathbf{p}_{i+2}$. Similarly, for Fig. 4(b) the shape is expanded by insertion of two slave points $\mathbf{s}_1^i$ and $\mathbf{s}_2^i$, calculated on $\frac{1}{3}(t_{i+1} - t_i)$ and $\frac{2}{3}(t_{i+1} - t_i)$ respectively.

### 2.3 Point distribution model

Let $\mathbf{X}_k = [x_1 \; y_1 \; x_2 \; y_2 \; \dots \; x_i \; y_i \; \dots \; x_m \; y_m]^T$ denote a shape that is normalized, expanded and aligned among $r$ number of shapes in $\mathfrak{I}$. We build the point distribution model by at first the determination of the mean shape $\mathbf{X}_\mu$:

$$(11) \quad \mathbf{X}_\mu = \frac{1}{r}\sum_{k=1}^{r}\mathbf{X}_k$$

for each $\mathbf{X}_k$ we get

$$(12) \quad d\mathbf{X}_k = \mathbf{X}_k - \mathbf{X}_\mu$$

and deduce the covariance matrix

$$(13) \quad \mathbf{S} = \frac{1}{r}\sum_{k=1}^{r}d\mathbf{X}_k d\mathbf{X}_k^T$$

from which we determine every eigenvectors $\mathbf{e}_l$ and eigenvalues $\lambda_l$, in which $\lambda_l$ is the $l$th eigenvalue of $\mathbf{S}$, $\lambda_l \geq \lambda_{l+1}$. Finally, we can approximate any shape [9] in $\mathfrak{I}$ using

$$(14) \quad \mathbf{Z} = \mathbf{X}_\mu + \mathbf{Fb}$$

where

$$(15) \quad \mathbf{F} = \begin{bmatrix} \mathbf{f}_1 & \mathbf{f}_2 & \dots & \mathbf{f}_l & \dots \mathbf{f}_g \end{bmatrix}$$

is the matrix of the first $g$ eigenvectors, and

$$(16) \quad \mathbf{b} = \begin{bmatrix} b_1 & b_2 & \dots & b_l & \dots b_g \end{bmatrix}^T$$

is the vector that controls the variation of the shape, with restriction [9] on each $b_l$ by

(17) $\qquad -2\sqrt{\lambda_l} \leq b_l \leq 2\sqrt{\lambda_l}$

so that the projected shape remains relevant to the samples in $\mathfrak{I}$. $g$ is deduced from a ratio $\phi$ such that

(18) $\qquad \dfrac{\sum_l^g \lambda_l}{\sum_l \lambda_l} > \phi$

Figure 5, 6, 7 and 8 illustrate the spline models for heart ventricle and hand. The models were reproduced from the examples printed in the paper by Cootes et al. For heart ventricle, we used 35 points to delineate the shape (see Fig. 5), whereas the original shape model employed 96 points. Similarly, for hand shape (Fig. 7) we used 48 points and the original shape model took 72 points.

The changes in shapes induced by the variation on $b_1, b_2, b_3$ are demonstrated in Figure 6 (heart ventricle) and Figure 8 (hand). The ways the shapes vary as a consequence of the changes on $b_1, b_2, b_3$ in general look similar to the original shape models, despite fewer landmark points are used in this study. Hence, it is safe to conclude that the nature of the original shape model is still well preserved under our formulation.

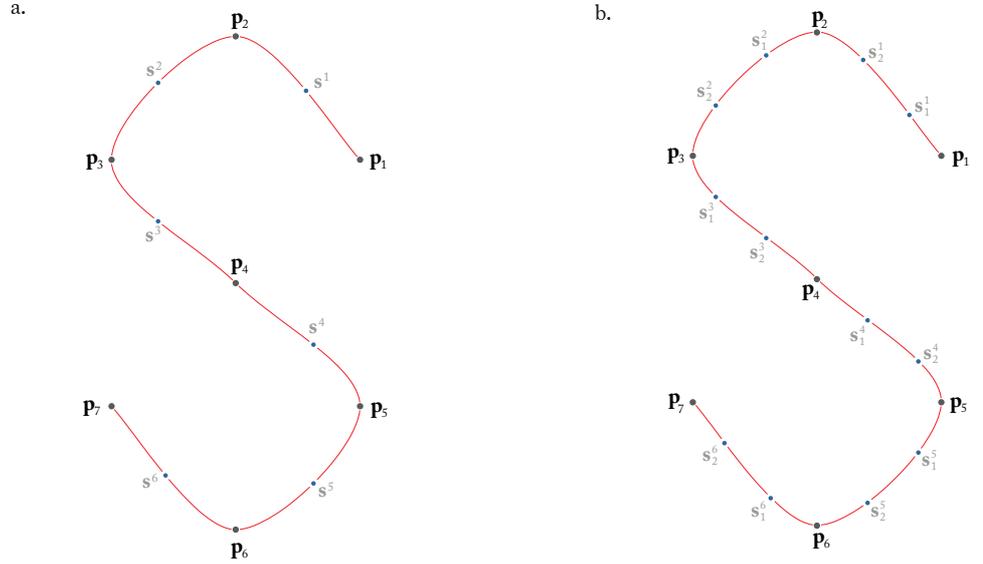

**Fig. 4.** Expansion of shape. (a) insertion of one slave point. (b) insertion of two slave points. For both (a) and (b), master points are colored in black and slave points in gray.

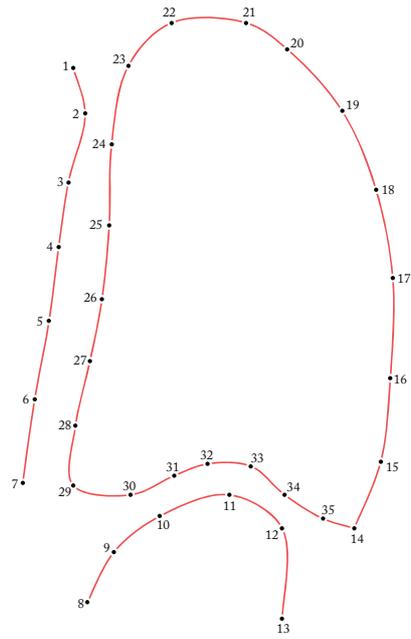

**Fig. 5.** Template for point distribution model of heart ventricle shape. Cootes et al. [9] used 96 points to draw the shape. We use only 35 points.

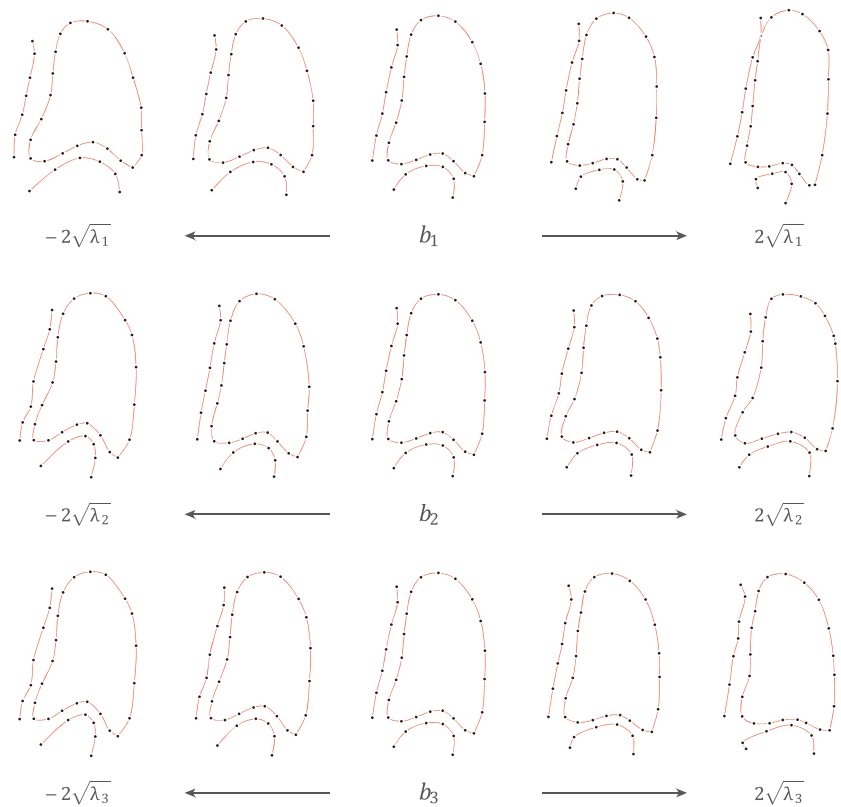

**Fig. 6.** Effects of individual variation in $b_1$, $b_2$, $b_3$ on the heart ventricle shape. The point distribution model comprises only master points, no slave point is inserted.

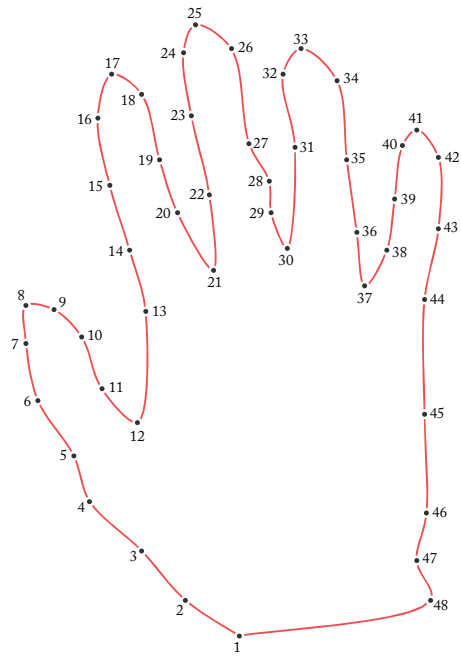

**Fig. 7.** Template for point distribution model of hand shape. Cootes et al. [9] used 72 points to draw the shape. We use 48 points.

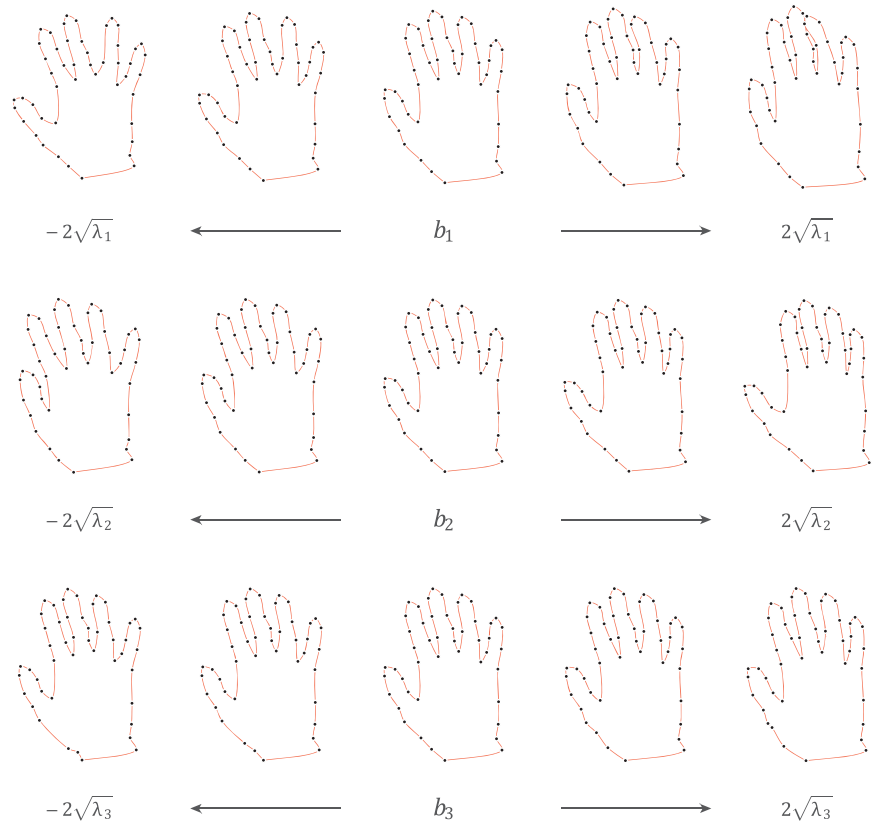

**Fig. 8.** Effects of individual variation in $b_1$, $b_2$, $b_3$ on the hand shape. In this model, for every master point a slave point is inserted after, except the last one (point 48). Black dots denote master points.

Let $\mathbf{Z}^n$ denote the shape produced by $\mathbf{b}^n$ through equation (14) at iteration $n$. The shape is normalized, thus each point in $\mathbf{Z}^n$ satisfies $0 \leq x, y \leq 1$. We say a shape is in *model scale* whenever a shape possesses the above property. For object segmentation, however, a shape must be in *image scale* for the action to happen. To do so, we transform $\mathbf{Z}^n$ into $\mathbf{Z}^{n+}$ with $\theta^n$, $s^n$, $\tau_x^n$, $\tau_y^n$ by equation (7), and subject the shape to an external force field. Let $\mathbf{v} = [u(x,y) \quad v(x,y)]$ of image $I(x,y)$ denote gradient vector flow field (GVF) [15], a dense vector field we use as the external force field that minimizes the below functional

$$\varepsilon = \iint \left[ \mu \left( u_x^2 + u_y^2 + v_x^2 + v_y^2 \right) + |\nabla f_e| \, |\mathbf{v} - \nabla f_e|^2 \right] \mathrm{d}x\mathrm{d}y \tag{19}$$

where $\mu$, a parameter that regulates the terms in the integrand, is advised to use higher value when image is noisy; $f_e$ is an edge map, which generally is set to $|\nabla I(x,y)|$. Since $\mathbf{Z} = [\mathbf{z}_1 \; \mathbf{z}_2 \; ... \; \mathbf{z}_i \; ... \; \mathbf{z}_m]^T$ and $\mathbf{z}_i = [x_i \; y_i]$, we define

$$\mathbf{q}_i^n = \mathbf{v}\left(\mathbf{z}_i^{n+}\right) \tag{20}$$

Let $\mathbf{d}_i^n$ denote the vector component of $\mathbf{q}_i^n$ in the direction perpendicular to the shape at $\mathbf{z}_i^n$ of iteration $n$ (see Fig. 9), then we have

$$\mathbf{Z}^{n\times} = \mathbf{Z}^{n+} + \Psi \mathbf{D}^n \tag{21}$$

where $\Psi$ is a parameter that adjusts the effect of GVF on shape and $\mathbf{Z}^{n\times}$ is the deformed $\mathbf{Z}^{n+}$. As $\mathbf{Z}^{n\times}$ is in image scale, to get $\mathbf{b}^{n+1}$, the updated vector that controls variation of the shape, we need to put the shape back again in model scale. Let $\mathbf{Z}^{n-}$ denote the model scale of $\mathbf{Z}^{n\times}$, transformed by $\theta^{n-}$, $s^{n-}$, $\tau_x^{n-}$, $\tau_y^{n-}$. Since $\mathbf{F}^{-1} = \mathbf{F}^T$ [9], we can determine $\mathbf{b}^{n+1}$ by

$$d\mathbf{Z}^n = \mathbf{Z}^{n-} - \mathbf{X}_\mu \tag{22}$$

and

$$\mathbf{b}^{n+1} = \mathbf{F}^T d\mathbf{Z}^n \tag{23}$$

At times it may be necessary to re-scale $\mathbf{b}$ as $\mathbf{Z}$ has strayed into a shape dissimilar to the intended object. We define

$$D_m^2 = \sum_{l=1}^{g} \left( \frac{b_l^2}{\lambda_l} \right) \leq D_{\max}^2 \tag{24}$$

and re-scale $\mathbf{b}$ when $D_m > D_{\max}$ by

$$b_l \leftarrow b_l \cdot \frac{D_{\max}}{D_m} \tag{25}$$

for $l = 1,2,...,g$. The above steps (20)—(25) are repeated until no significant change is observed between $\mathbf{Z}^n$ and $\mathbf{Z}^{n+1}$.

For now the general procedures (see also Fig. 10) to update a shape in search of object of interest are presented, but we are yet to detail the steps to get $\theta^n$, $s^n$, $\tau_x^n$, $\tau_y^n$ and $\theta^{n-}$, $s^{n-}$, $\tau_x^{n-}$, $\tau_y^{n-}$. When $n$ equals to 1, the first iteration, $\theta^1$, $s^1$, $\tau_x^1$, $\tau_y^1$ are either user-defined or pre-determined by algorithm that handles initialization, after that, then

(26) $\qquad \theta^n, s^n, \tau_x^n, \tau_y^n \xleftarrow{\text{eq. (8)}} \mathbf{Z}^{(n-1)\times}, \mathbf{Z}^n$

On the other hand, to determine $\theta^{n-}$, $s^{n-}$, $\tau_x^{n-}$, $\tau_y^{n-}$, we need to first define

(27) $\qquad \delta^n = \mathbf{Z}^{n+} - \mathbf{X}_\mu^{n+}$

where $\mathbf{X}_\mu^{n+}$ is the $\mathbf{X}_\mu$ in image scale transformed by $\theta^n$, $s^n$, $\tau_x^n$, $\tau_y^n$, and we get

(28) $\qquad \theta^{n-}, s^{n-}, \tau_x^{n-}, \tau_y^{n-} \xleftarrow{\text{eq. (8)}} \left( \mathbf{Z}^{n\times} - \delta^n \right), \mathbf{X}_\mu$

For the first iteration, however, we determine $\delta^1$ by

(29) $\qquad \delta^{1-} = \mathbf{Z}^1 - \mathbf{X}_\mu$

(30) $\qquad \delta^1 \xleftarrow{\text{eq. (7)}} \delta^{1-}, \theta^1, s^1, \tau_x = 0, \tau_y = 0$

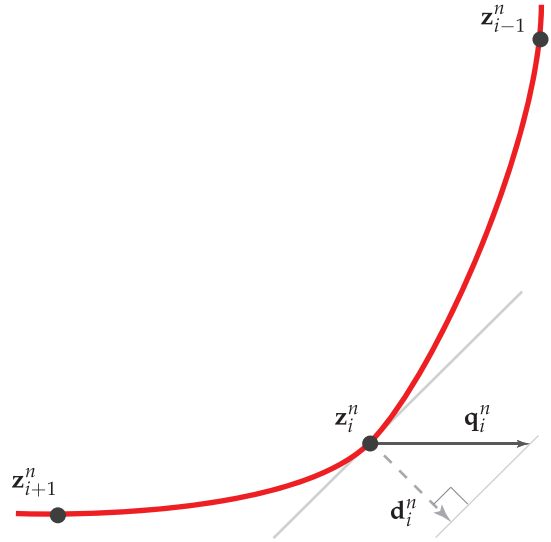

**Fig. 9.** The displacement vector $\mathbf{d}_i^n$ at point $\mathbf{z}_i^n$.

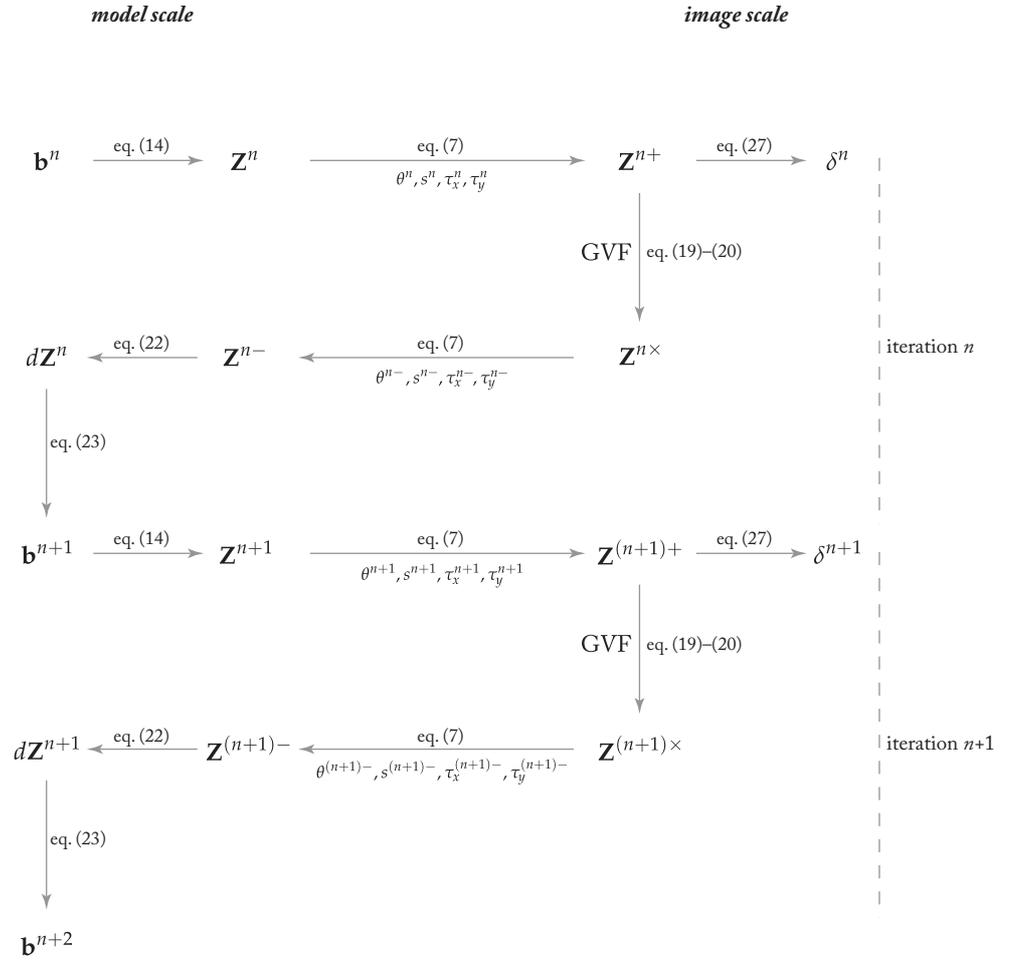

**Fig. 10.** Segmentation steps at iteration *n* and *n+1*. This illustration focuses on the evolution of **b**ⁿ to **b**ⁿ⁺², with the shape going back and forth between model scale and image scale.

### 2.5 Edit

After the segmentation, edit of contour is simply achieved by shifting of master points (control points). Figure 11 shows an example of the process. As illustrated, there is a section of curve lying low and flat, controlled by three master points. Suppose the entire contour is desired to be a sinusoidal-like curve, with peaks almost at the same height (outlined in green line), what is necessary to get the shape then, is just a simple shift of the master point (green dot) that sits on the lower peak to a higher position. In the process only the section that involves the shifted point is changed, the rest remains mostly the same as before.

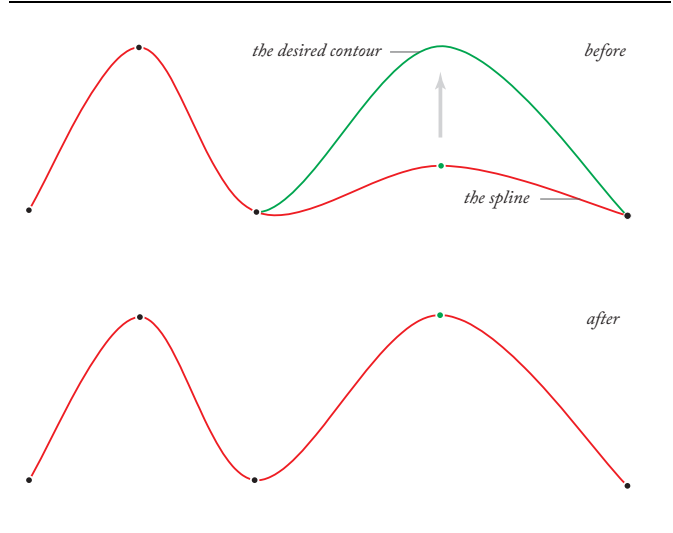

**Fig. 11.** Edit of contour. The master point on the lower peak is moved to a higher position in order to form the desired shape.

*2.6 Evaluation*

When ground truth is available, we can always measure the accuracy of a segmentation by overlap measure [19]:

$$\Theta = \frac{\text{TP}}{\text{TP} + \text{FP} + \text{FN}}$$

(31)

where TP is the sum of the pixels correctly labeled as part of the objects, FP the sum of the pixels incorrectly labeled as the objects, and FN the number of pixels belonged to the objects but identified as the otherwise. It is important to note that, however, when the ground truth is delineated by polygon, the overlap measure for segmentation by spline can never reach value of 1 even when the smooth boundary is a better delineation.

To evaluate the manual edit after segmentation, we first define *manipulation* as

$$\text{M} = \frac{\partial}{\aleph}$$

(32)

where $\partial$ is the duration to complete the edit, $\aleph$ is the number of actions carried out within the time frame. The lower the number, the easier the handling of the edit. Besides, we also define *effort*

$$\text{E} = \text{M} \times \partial = \frac{\partial^2}{\aleph}$$

(33)

which tells how much handling is necessary for each segmentation. Finally we have efficiency

$$\text{Y} = \frac{\Delta\Theta \times 100}{\text{E}}$$

(34)

where $\Delta\Theta$ is the change in overlap measure before and after the manual edit.

## 3. Results

*3.1 Ocular thermography*

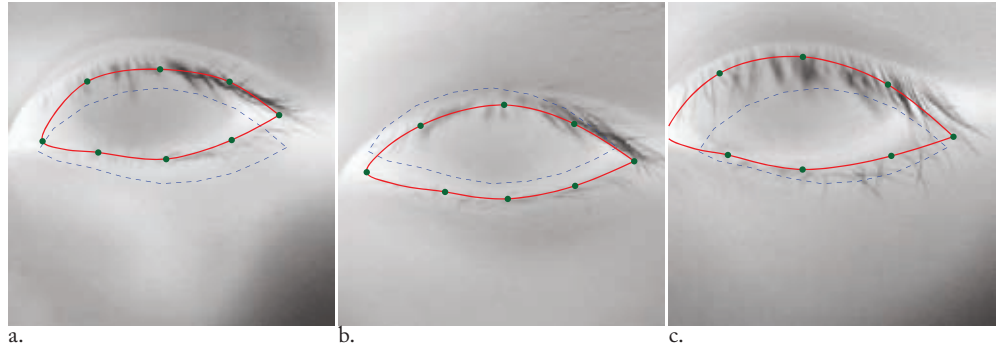

**Fig. 12.** Segmentation on ocular thermographs. Red lines denote the final contour, with green dots represent the master points at final position. Blue dotted line marks the contour of initial instance.

Fig. 12 illustrates the application of our method on the ocular thermographs (left eye only) collected at [20]. The point distribution model was formulated on a sample of 12 eyes, with each eye delineated by 8 master points, and one slave point was inserted every after master point. We set $\phi$ to be 0.95, which by equations (18) gave $g = 6$. The edge map is

$$(35) \qquad f_e = |\nabla I_m(x, y)|$$

where $I_m$ is an image convoluted with a median filter [21]. The size of the median filter kernel is 4x4.

The initial conditions for segmentations are fixed: 0, 105, 120, 110 for $\theta^1$, $s^1$, $\tau_x^1$, $\tau_y^1$ respectively; 250 iterations were evaluated to deliver the segments. $D_{\max}$ was set at $2g = 12$ to control the **b**.

*3.2 Fundus image*

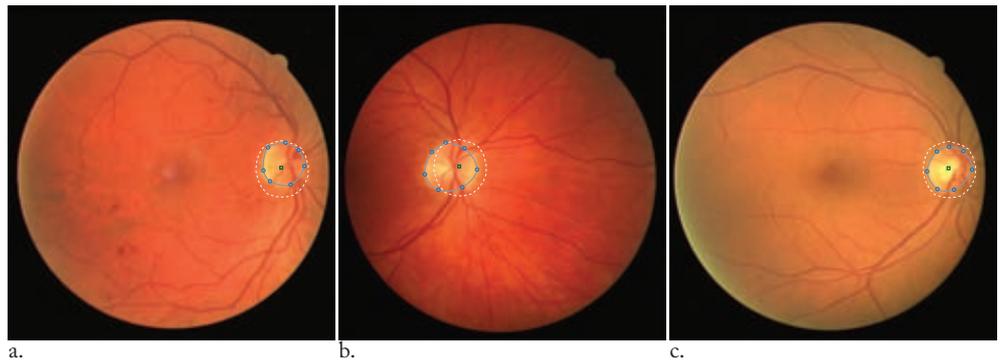

**Fig. 13.** Segmentation of optic disc on fundus images. Blue lines denote the final contour, with circular dots represent the master points at final position. White dotted line marks the contour of initial instance.

Fig. 13 shows the segmentation of optic disc on the fundus images of DRIVE dataset [22]. 17 images were used to formulate the point distribution model, in which each shape of optic disc comprises 7 master points and 2 slave points every after master point. $\phi$ was set to 0.95, and that gave g = 6, with $D_{\max}$ put at $2g = 12$. The edge map

$$(36) \quad f_e = \left| \nabla I_{m,b}(x,y) \right|$$

is used, where $I_{m,b}$ is the blue component of the image, convoluted with a 2x2 median filter.

In the above examples, we used the method proposed by [23] to determine the center of optic disc, and scaled up the model 50 times to initiate the first instance. 450 iterations were evaluated to produce the above segmentations.

### 3.3 Chest radiograph

### 3.3.1 Automated segmentation

Unlike previous examples, for chest radiographs we performed automated lung segmentation solely using active spline model. The images (see Table 1) in used were taken from the database put public by [24], and split into two folds: one fold comprised the odd numbered images, and the other even numbered radiographs. On each fold we built a point distribution model, and performed segmentation on the other fold where the model was not based on, so that all the images were tested for segmentation.

Fig. 14 shows the point distribution model of the lungs, and its variations on $b_1$, $b_2$, $b_3$ is illustrated in Fig. 15. We run the segmentation on a standard image pyramid [25] comprising 5 levels of resolution in order to overcome the conundrum of initialization. Table 2 tabulates the details of the setup in the segmentation.

**Table 1**
The JSRT database [24]. Each 'nodule' image in the data set contains exactly one lung nodule.

| Type | No. |
| --- | --- |
| nodule | 154 |
| non-nodule | 93 |
| Total | 247 |

**Table 2**
Details of the lung automated segmentation. The edge map is $|\nabla I(x,y)|$ if median filter is not used, otherwise we use equation (35) to generate the edge map with the kernel size specified in the table. The segmentation starts at level 5 and ends at level 1.

| Level | Resolution | Size of median filter kernel | No. of iteration ($q$) | $c_t$ | $c_s$ | $c_x$ | $c_y$ | $c_b$ | $c_{b2}$ | $\phi$ |
| --- | --- | --- | --- | --- | --- | --- | --- | --- | --- | --- |
| 5 | 16 X 16 | nil. | 1 | nil. | nil. | nil. | nil. | nil. | −3.0 | 0.50 |
| 4 | 32 X 32 | nil. | 2 | 0.2 | 2.0 | 2.0 | 2.0 | 0.5 | −1.5 | 0.60 |
| 3 | 64 X 64 | 3 X 3 | 10 | 0.2 | 2.0 | 2.0 | 2.0 | 0.5 | −1.5 | 0.70 |
| 2 | 128 X 128 | 4 X 4 | 60 | 0.2 | 2.0 | 2.0 | 2.0 | 0.5 | −1.5 | 0.90 |
| 1 | 256 X 256 | nil. | 180 | 0.2 | 2.0 | 2.0 | 2.0 | 0.5 | −1.5 | 0.98 |

**Table 3**

Overlap values for lung segmentation. We use 'ASPLM' to denote the automated segmentation detailed in section 3.3.1, 'ASPLM adjusted' for the spline model with manual edit (see section 3.3.2). The results of 'PC post-processed', 'ASM tuned', 'AAM whiskers BFGS', 'AAM whiskers', 'ASM default' and 'AAM default' were determined by Ginneken et al [19].

| Methods | Average | Min | Max |
|---|---|---|---|
| ASPLM adjusted | 0.945±0.008 | 0.925 | 0.969 |
| PC post-processed | 0.945±0.022 | 0.823 | 0.972 |
| ASM tuned | 0.927±0.032 | 0.745 | 0.964 |
| AAM whiskers BFGS | 0.922±0.029 | 0.718 | 0.961 |
| AAM whiskers | 0.913±0.032 | 0.754 | 0.958 |
| ASM default | 0.903±0.057 | 0.601 | 0.960 |
| ASPLM | 0.879±0.047 | 0.588 | 0.937 |
| AAM default | 0.847±0.095 | 0.017 | 0.956 |

The automated segmentation begins at level 5, where the image size is 16x16. Parameters $\theta$, $s$, $\tau_x$, $\tau_y$ were set to 0, 6, 9, 7 respectively for the algorithm to begin, and $b_l$ is set to 0, except $b_2$ is put at $-3\sqrt{\lambda_2}$. The lung shape as a result is long and slim, which in overall gives better segmentation outcome.

For simpler illustration, we further define $^n\theta^k$, $^ns^k$, $^n\tau_x^k$, $^n\tau_y^k$ and $^n\mathbf{b}^k = \begin{bmatrix} ^nb_1^k & ^nb_2^k & ... & ^nb_l^k & ... & ^nb_g^k \end{bmatrix}^T$ as the parameters $\theta$, $s$, $\tau_x$, $\tau_y$ and $\mathbf{b}$ at iteration $k$ of level $n$. Assume at level $j$ we preset $q$ number iteration to run, then the initial values for $\theta$, $s$, $\tau_x$, $\tau_y$ and $b_l$ of level $j-1$ are $c_t$, $c_s$, $c_x$, $c_y$, and $c_b$ times $^j\theta^{q+1}$, $^js^{q+1}$, $^j\tau_x^{q+1}$, $^j\tau_y^{q+1}$ and $^jb_l^{q+1}$ respectively, where $^j\theta^{q+1}$, $^js^{q+1}$, $^j\tau_x^{q+1}$, $^j\tau_y^{q+1}$ are determined from $\mathbf{Z}^{q\times}$, $\mathbf{Z}^{q+1}$ of level $j$ by equation (26); $^jb_l^{q+1}$ is obtained by equation (23). However, we do not apply $c_b$ on $b_2$, instead the initial value for $b_2$ of level $j-1$ is set as $c_{b2}\times\sqrt{\lambda_2}$.

Hence, for example, the initial values for $\theta$, $s$, $\tau_x$, $\tau_y$ and $b_l$ of level 2 are 0.2, 2, 2, 2 and 0.5 times the $^3\theta^{11}$, $^3s^{11}$, $^3\tau_x^{11}$, $^3\tau_y^{11}$ and $^3b_l^{11}$ respectively; whereas $b_2$ is equaled to $-1.5\times\sqrt{\lambda_2}$ (see Table 2). $D_{\max}$ is $2g$ throughout all levels.

Fig. 16 illustrates some of the segmentation results, and Table 3 tabulates our findings. We used the ground truth put online by Ginneken et al [19]. and achieved in average an overlap of 0.879 for the entire data set (247 images), with a median, minimum, maximum of 0.890, 0.588, 0.937 respectively. There were 89 images in which the segmentation gave an overlap value above 0.90, and each segmentation in the data set took 3.63s in MATLAB R2012b to complete. We run the trial on a i5 2.5GHz Intel Core processor with 4GB 1600MHz DDR3 Ram.

In Table 3 we have compared our results against several other methodologies for references, all performed by Ginneken et al [19]. The "ASM default" and "AAM default" segment lung by the original active shape model and active appearance model. "ASM tuned" uses active shape model with parameters tuned to the image modality; "AAM whiskers" is active appearance model with boundary information added implicitly. On the other hand, "AAM whiskers BFGS" is "AAM whiskers" optimized by a quasi-Newton method, whereas "PC post-processed" is pixel classification with some post processing to rectify segment outcome.

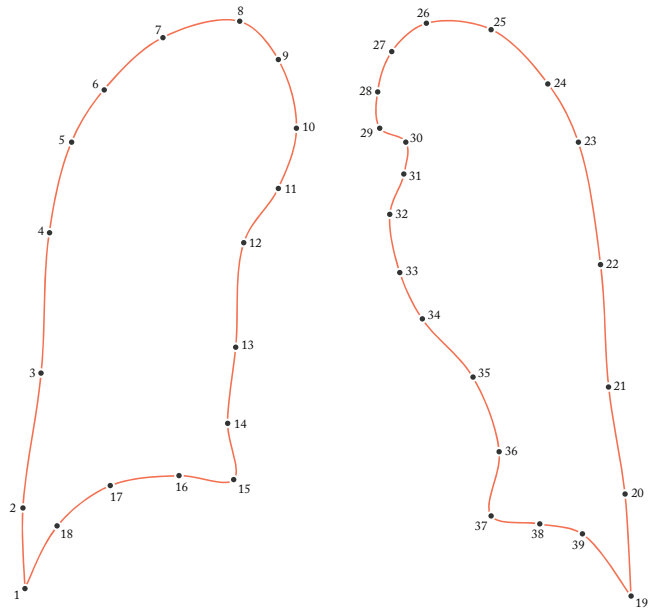

**Fig. 14.** Template for point distribution model of lungs. We use only 18 points to draw right lung, and another 21 for left lung. Whereas Ginneken et al. [19] used 44 points for right lung, and 50 points for the left.

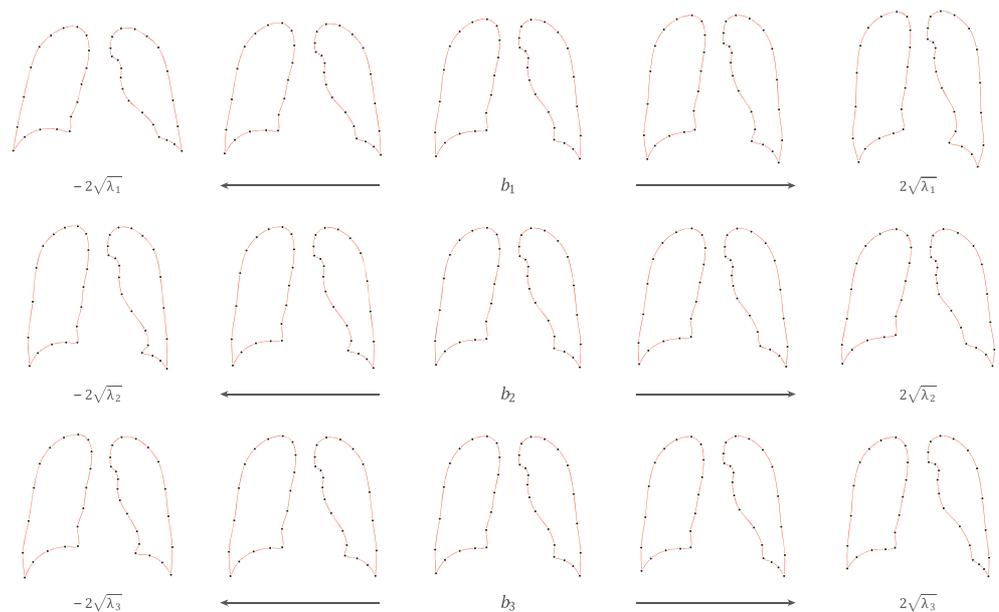

**Fig. 15.** Effects of individual variation in $b_1$, $b_2$, $b_3$ on the lung shape. In this model, for every master point two slave points are inserted after. Black dots denote master points.

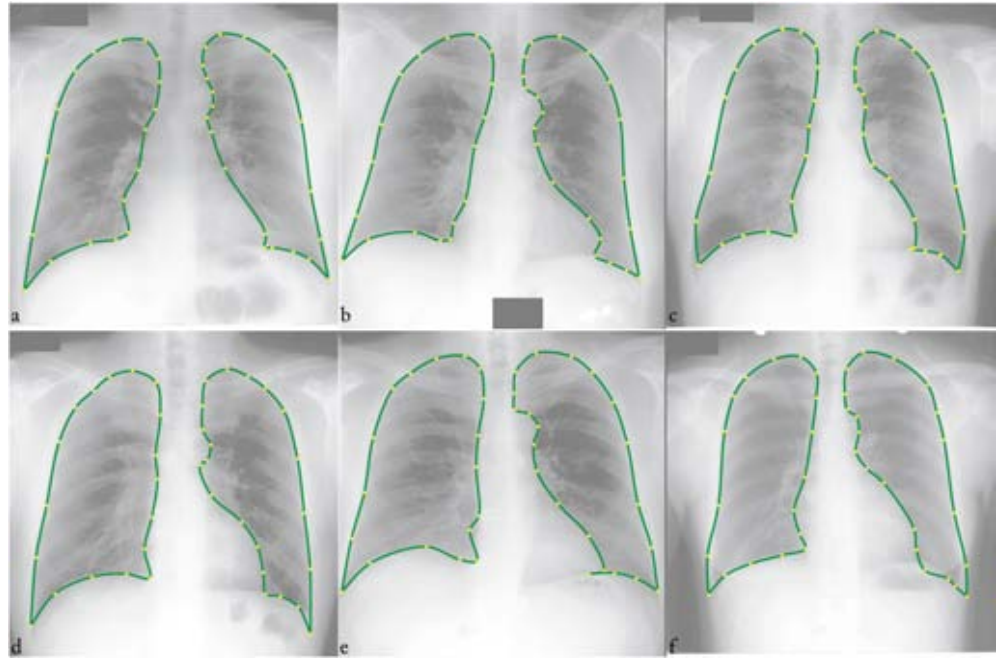

**Fig. 16.** Segmentation outcomes on some of the radiographs. The overlap values for each image are: (a) 0.9172  (b) 0.9074  (c) 0.9161  (d) 0.9269  (e) 0.9242  (f) 0.9135. Yellow dots denote master points.

*3.3.2 Manual Edit*

Fig. 17 shows the edit of a segment. The concern of the edit was at the bottom corner of the right lung, next to the right thoracic wall and diaphragm. Two master points were shifted, a re-evaluation of spline ensued, and the desired contour was delivered in just two moves.

To further evaluate the manual edit, we have developed a procedure to study the process. As illustrated in Fig. 18a, we at first loaded a radiograph (image size of 512x512) and ran the automated segmentation. When the segment was ready, the algorithm immediately displayed the same radiograph but with its gray-level inverted and ground truth superimposed (see Fig. 18b), so that the user could proceed to edit and export the final segment with ground truth in view (hence radiologist was not necessary for the study). In the process of the edit he was not informed of the overlap values before and after the procedure, to ensure the judgment was not interfered. The automated segmentation used only the model trained on odd-numbered images, although the experiment ran through all the 247 images.

The automated segmentation in this case gave in average an overlap of 0.877, and after the manual intervention, it went up to 0.945. We define the number of moves executed to shift the points in an edit as ℵ, the number of actions. In average, 15.7 ± 5.2 moves were used within a mean duration of 30.4 ± 10.1s to improve the delineation (see Table 4 for more details). Fig. 19 shows the efficiency of each edit against the overlap value before the edit.

**Table 4**

The study of manual edit by 1 tester on all 247 images.

|  | Average | Min | Max |
|---|---|---|---|
| ASPLM (overlap, after edit) | 0.945±0.008 | 0.925 | 0.969 |
| ASPLM (overlap, before edit) | 0.877±0.048 | 0.588 | 0.937 |
| Duration (second) | 30.4±10.1 | 4.6 | 72.1 |
| Number of moves | 15.7±5.20 | 3 | 36 |
| Efficiency | 0.114±0.060 | 0.010 | 0.370 |

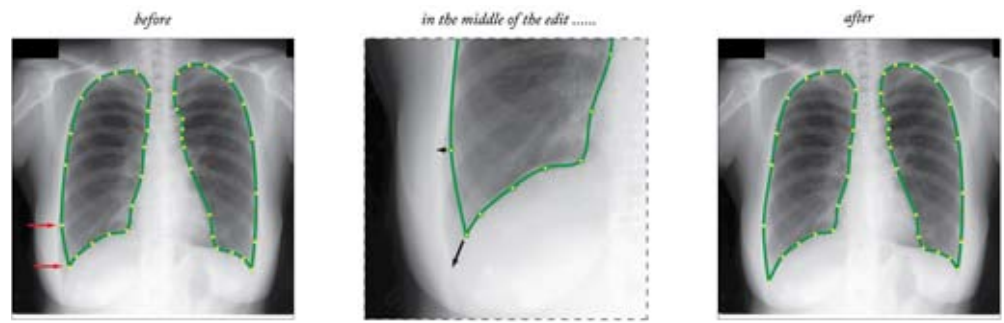

**Fig. 17.** The edit of a segment, at the lower corner of the right lung, by a shift of two master points.

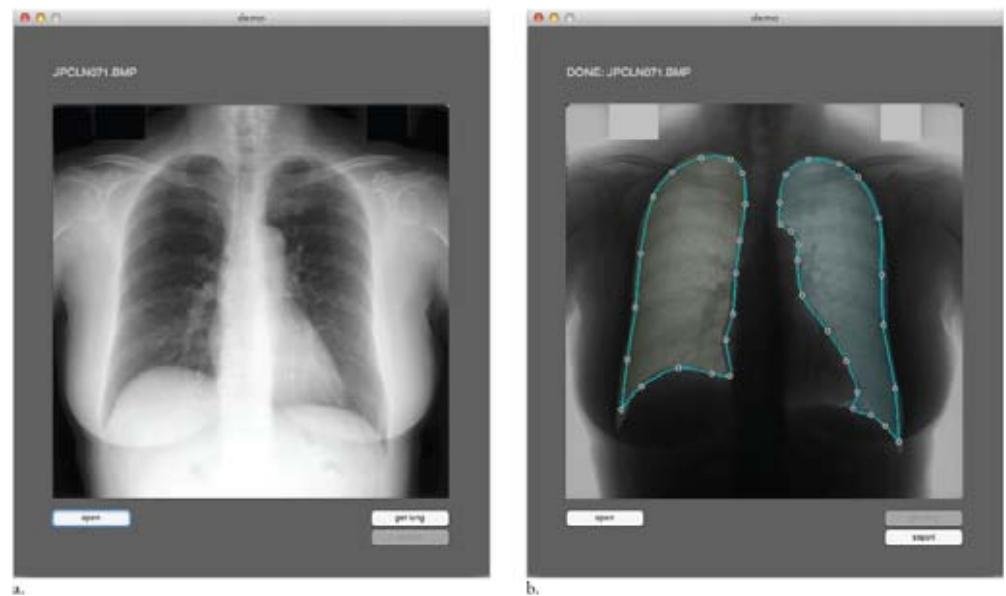

**Fig. 18.** Setup for the study of manual edit. (a) before automated segmentation (b) after automated segmentation, with gray-level inverted and ground truth superimposed (the ground truth is coloured).

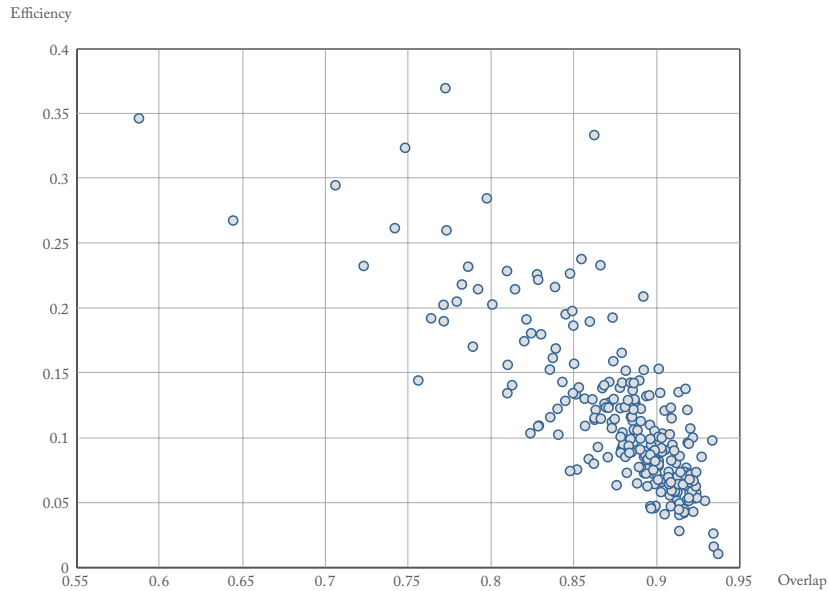

**Fig. 19.** Efficiency of edits versus overlap values. We plot the overlap value of each automated lung segmentation on x-axis (i.e. before manual intervention), and the efficiency of the subsequent edit on y axis.

## 4. Discussion

Liu and Udupa stated five limitations of active shape models in their seminal paper for Oriented Active Shape Models [26]. According to their experience, the method is sensitive to search range and initialization; it requires numerous landmarks, and the delineation accuracy is lower. Furthermore, the method is unable to fully exploit specific information present in an image. Instead, they asserted that the approach they proposed, which fused active shape models and live wire [27], overcome the above issues.

Their method did circumvent the conundrum of initialization, but was achieved by an extensive search over image for the placement of shape instance with the use of a cost function, *without* consideration of scale and orientation. And that capability is an add-on to the segmentation strategy, not a characteristic *inherent* in the algorithm core. They reported improvement in accuracy by live wire, but of puny significance to some users. The noisy boundary is sometimes undesirable, and so is the lost of shape model.

However, we agree that the amount of landmarks required for the development of active shape models is burdensome, and thus incorporate CRS into point distribution model in this study to alleviate the issue. And such a fusion yields true shape model, that is, a shape model independent of scale, orientation and position.

Consider the edit of lung segment as illustrated in Fig. 17. If that were to be implemented in reality, the image must be displayed at the original scale (around 2048 x 2048) and the curve be substantially scaled up, so that radiologist could manipulate the edit as per normal. That poses no issue for active spline models, but for other methods their boundary will be rough and abrupt.

Therefore unlike Liu and Udupa, our primary concern in regard to active shape models is its usability during the formation of the model and the manual intervention after a segmentation. We put more emphasis on the interactive part of the method instead of the computational part, contrary to the majority of literature.

According to Olabarriaga and Smeulders [8], the usability of an interactive strategy for segmentation can be qualitatively evaluated against three aspects. Accuracy refers to the capability of the method to segment any desired region. Repeatability refers to the variation in delineations that arises when the strategy is performed at various session on a single image. The last is efficiency, which is further broken down into four components: the elapsed time to deliver a delineation or an update, the demand on mouse operation, the necessary knowledge to operate the tool, and the predictability in response to user input.

With these one can see the problems in corrective measures that proceeds edit at pixel level, or on the numerous points that encircles an object. These measures can no doubt achieve the highest accuracy, but since they perform the edit at the lowest level, they also necessitates the highest exactitude in operation. The user gets fatigue easily, and the operation then becomes sluggish.

A good interactive strategy for segmentation should demand as few mouse clicks and attention as possible. Simple manipulation of parameters is one of the solutions, but it poses another snag: the operator is forced to work through the task in two disparate knowledge domains. On one hand the operator sees the delineation in the domain of the imaging modality, but to manipulate the parameters he needs to consider the problem in the domain of the algorithm. This demands the user to have considerable understanding on the method, though reasonable, it remains an obstacle to the adoptability of the method. Even if the user is willing to learn, in actual practice tardy operation can easily occur.

Active spline model in contrast demands only simple mouse operations, in which each operation is almost a no-brainer. The update of curve is fast, and the outcome highly repeatable. The centripetal parameterization of CRS makes the changes in curve very predictable, since the curve always tightly follows the points with no cusp or intersection, as elucidated in Fig. 2, 3 and 11.

On the other hand, by parameters $\aleph$, M, E, $\Upsilon$ we have quantitatively evaluated the edit operation, albeit incompletely. In average it took about 1.94s to execute an action, as determined by M, the manipulation. Though not a solid proof, it at least shows the edit is easy to operate. The averaged duration needed for a complete edit was 30.4s, as no similar study was conducted on the dataset, we have no basis to make comparison. However, as a reference, the PC post-processed (see Table 3) by Ginneken et al. [19] took about 30s to complete the necessary segmentation using computers of the time the paper published.

We used only 1 tester in the edit study, intended only to give a rough idea, not as a rigorous investigation. The reason of this plight is simple: it is hard to get volunteer willing to go through all the 247 images. Although one may argue it is not necessary for tester to edit all of them, Fig. 19 probably tells the otherwise.

For an automated segmentation that almost encircles the desired regions (high overlap value), the room for improvement in this case is marginal, but to realize the improvement considerable effort is still required (according to our experiences). Often the effort necessary in this case is comparable to the amount of effort put in on a much more inferior delineation by the automated segmentation. If the effort demanded is similar, yet the improvement in accuracy is small, the efficiency as a result is always lower (by equation (34)). This explains the inverse relationship between efficiency and the overlap value of the automated segmentation, as demonstrated in Fig. 19. And hence higher accuracy in automated segmentation is not always beneficial to the user from the practice point of view, because the effort required to make the improvement is generally not lesser. Even if a segmentation is almost perfect, intervention may still be necessary at various circumstances. Therefore for segmentation we should focus more on the intervention after if we truly want to improve the usability of computing method.

The advantage of ASPLM is more apparent when compared with Bézier Cubics [28]. As illustrated in Fig. 20a, for Bézier Cubics, it takes two control points $p_A$, $p_b$ to vary a curve segment AB. But Kang et al. [29] thought it was inconvenient and tough to operate two control points simultaneously, and thus proposed to combine $p_A$, $p_b$ into $p_c$, so that user needs only to manipulate one control point in their proposed editing tool (see Fig. 20b). Nevertheless, the handling is less than intuitive in comparison to our solution, where the curve goes through control points (see $p_d$ in Fig. 20c) and thus the operation is much more straightforward.

The disadvantage of ASPLM, on the other hand, is that the room for fine edit is rather restrained, since curves are determined only by a handful of points. Addition of points to shape or model can help, yet the subsequent efficiency may suffer. We can have either excellent efficiency or accuracy, but not the both.

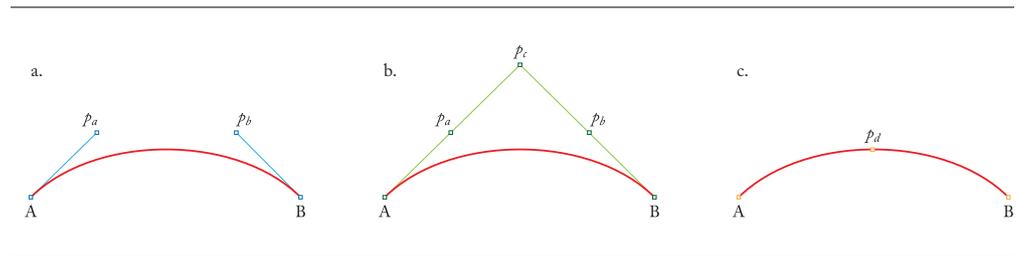

**Fig. 20.** Curve segment AB generated by (a) Bézier Cubics (b) method by Kang et al., where $p_A$ falls in the middle between A and $p_c$, $p_b$ in the middle between B and $p_c$ (c) centripetal parameterized Catmull-Rom spline.

Lastly, although we used only biomedical images as examples, the algorithm is equally applicable to other fields. In applications where active shape model or active appearance model is already used, our solution should work fine. And automated segmentation is always achievable for ASPLM if initialization is properly handled, as demonstrated in 3.2 and 3.3. With these not only we have a complete solution, but also an approach that provides easy edit on segment.

## 5. Conclusion

In this article we have fused point distribution model and centripetal-parameterized Catmull-Rom spline into a new active segmentation strategy. The method, termed active spline model, uses fewer points to describe shape yet draws smoother curve on any scale, delivers both model-based segmentation and a measure to edit the segment afterward. Gradient vector flow field was employed to drive the deformation in shape instance. The edit on segment is achieved by shift of points to update the curve.